\documentclass[pdflatex,sn-mathphys-num]{sn-jnl}

\usepackage{graphicx}%
\usepackage{multirow}%
\usepackage{amsmath,amssymb,amsfonts}%
\usepackage{amsthm}%
\usepackage{mathrsfs}%
\usepackage[title]{appendix}%
\usepackage{xcolor}%
\usepackage{textcomp}%
\usepackage{manyfoot}%
\usepackage{booktabs}%
\usepackage{algorithm}%
\usepackage{algorithmicx}%
\usepackage{algpseudocode}%
\usepackage{listings}%

\usepackage{tcolorbox}
\usepackage{array}    
\usepackage{ragged2e} 

\begin{document}

\title[The simulation of judgment in LLMs]{The simulation of judgment in LLMs}


\author[a]{Edoardo Loru}
\author[b]{Jacopo Nudo}
\author[c]{Niccolò Di Marco}
\author[d]{Alessandro Santirocchi}
\author[d]{Roberto Atzeni}
\author[b]{Matteo Cinelli}
\author[d]{Vincenzo Cestari}
\author[d]{Clelia Rossi-Arnaud}
\author[b,*]{Walter Quattrociocchi}

\affil[a]{Department of Computer, Control and Management Engineering, Sapienza University of Rome}
\affil[b]{Department of Computer Science, Sapienza University of Rome}
\affil[c]{Department of Legal, Social, and Educational Sciences, Tuscia University}
\affil[d]{Department of Psychology, Sapienza University of Rome}

\affil[*]{walter.quattrociocchi@uniroma1.it}


\abstract{\textbf{\textcolor{red}{Please refer to published version:}} \url{https://doi.org/10.1073/pnas.2518443122}\\Large Language Models (LLMs) are increasingly embedded in evaluative processes, from information filtering to assessing and addressing knowledge gaps through explanation and credibility judgments. This raises the need to examine how such evaluations are built, what assumptions they rely on, and how their strategies diverge from those of humans. We benchmark six LLMs against expert ratings—NewsGuard and Media Bias/Fact Check—and against human judgments collected through a controlled experiment. We use news domains purely as a controlled benchmark for evaluative tasks, focusing on the underlying mechanisms rather than on news classification per se. To enable direct comparison, we implement a structured agentic framework in which both models and nonexpert participants follow the same evaluation procedure: selecting criteria, retrieving content, and producing justifications. Despite output alignment, our findings show consistent differences in the observable criteria guiding model evaluations, suggesting that lexical associations and statistical priors could influence evaluations in ways that differ from contextual reasoning. This reliance is associated with systematic effects: political asymmetries and a tendency to confuse linguistic form with epistemic reliability—a dynamic we term epistemia, the illusion of knowledge that emerges when surface plausibility replaces verification. Indeed, delegating judgment to such systems may affect the heuristics underlying evaluative processes, suggesting a shift from normative reasoning toward pattern-based approximation and raising open questions about the role of LLMs in evaluative processes.
}

\maketitle

\section*{Introduction}\label{sec1}
Large Language Models (LLMs) are increasingly embedded in workflows involving classification, evaluation, recommendation, and decision support \cite{binz2025should}. 
This rapid integration of AI technologies is not just reshaping industries but also presenting a fundamental choice between a path of pure automation and one of human complementation, where technology is designed to augment human spread and mitigate widespread unemployment \cite{zer2024_2}.
Beyond assistive tools, they may influence institutional decision-making, where automated outputs support high-stakes judgments. Advances like chain-of-thought prompting \cite{gao2025take} and semi-autonomous agents \cite{yax2024studying,wang2024survey} mark a broader shift: we are no longer just automating tasks, but embedding evaluative functions into socio-technical systems \cite{torkamaan2024challenges}.
This delegation carries significant risks, as AI systems trained on biased historical data may replicate and amplify societal inequalities in critical domains, such as the labor market \cite{zer2024_1}.

As these systems scale, the issue is no longer only whether outputs are correct, but how the very notion of judgment is operationalized once decisions are delegated to statistical models.
This raises a key question: what heuristics are encoded when decisions are delegated to LLMs, and how are classifications produced, justified, and interpreted?
LLMs can produce outputs similar to those of humans in structured tasks \cite{binz2025foundation}, but the similarity concerns results, not the process.
What appears as alignment at the output level may conceal a deeper epistemic shift, where normative reasoning is replaced by surface-level approximation.

We analyze how LLMs apply the concepts of reliability and bias, normative categories that influence which content is shown, hidden, or ignored. These classifications shape information exposure, platform moderation, and public trust. Understanding how models handle these constructs is necessary to assess their societal and epistemic impact.
To investigate this, we provide a benchmark involving six LLMs, two expert rating systems, and a sample of human evaluators. News outlets are classified by reliability and political bias, allowing us to compare both outputs and the procedures behind them. 
The news domain serves here only as a controlled testbed, enabling us to isolate and analyze the mechanisms behind automated evaluations rather than focusing on domain-specific accuracy.
This helps show how automated evaluations change not just efficiency, but also the criteria and assumptions used in decision-making.
Indeed, LLMs often use surface-level text patterns instead of reasoning from evidence \cite{qu2024performance,stureborg2024large}.

In online spaces where information spreads fast and users are split into echo chambers \cite{cinelli2021echo,holton2015reciprocity,khan2017social,avalle2024persistent,kubin2021role}, judging source credibility as well as their biases is a core problem \cite{budak2024misunderstanding,lazer2018science,del2016spreading}. These judgments shape what people believe and how public debate evolves—driving polarization, spreading misinformation, and eroding trust \cite{gallup2018indicators,newman2018digital,bail2018exposure,falkenberg2022growing,cinelli2020covid}.
Humans tend to rely on basic standards: is it accurate, independent, and transparent? \cite{metzger2015psychological,rieh2010credibility,metzger2010social}
Expert human evaluators—like NewsGuard and MBFC—apply clear rules to rate thousands of news sites \cite{luhring2025best,nakov2021survey,newsguard2024}. 
Our goal is not to assess whether LLMs can replace human raters, but to use these benchmarks to analyze the heuristics guiding their evaluations and how they operationalize concepts such as credibility and bias. Widely used models such as GPT \cite{openai2024}, Gemini \cite{team2024gemini}, and Llama \cite{meta2024}—already employed in numerous classification and fact-checking tasks \cite{tornberg2024large,gilardi2023chatgpt,wu2023large,chiang2023can,krugmann2024sentiment,yang2023large,hoes2023leveraging,quelle2024perils,hernandes2024llms}—offer a natural testbed for this analysis.
Recent studies have explored whether LLMs replicate human heuristics \cite{hu2024generative,yax2024studying,motoki2025assessing} or reflect ideological biases learned during training \cite{strachan2024testing,coppolillo2025unmasking,safdari2023personality}, but most focus on output-level metrics such as accuracy or bias. Few examine how these outputs are produced.
We address this gap by analyzing how LLMs generate judgments about reliability and political bias, and how their procedures compare to human evaluation.
Six LLMs classify 2,286 domains by reliability and political leaning. Their outputs are compared to expert ratings, and we examine the lexical patterns and explanatory cues they provide to infer the underlying heuristics.
We do not assume human-like reasoning. Rather, we empirically analyze how models operationalize evaluative tasks, allowing us to test whether their decisions rely on statistical associations shaped by training and prompting.
This leads us to ask whether delegating judgment to LLMs preserves its normative and epistemic meaning, or whether it transforms it into what we call epistemia—a condition in which the appearance of coherent and authoritative judgment arises from statistical patterning alone, producing the illusion of knowledge when surface plausibility substitutes for evidence-based reasoning.

Our findings show that model outputs often align with expert ratings of reliability and bias, yet systematic asymmetries emerge across the political spectrum. Moreover, LLMs generate consistent linguistic markers when explaining their evaluations. 
To test these differences, we implement a structured protocol in which LLMs simulate evaluative behavior—selecting criteria from a predefined set, retrieving content, and producing justifications—while human participants follow the same procedure in a controlled setting. The results show that LLMs and humans prioritize different reliability criteria, consistent with a shift from context-dependent, normative reasoning—understood here as the application of explicit quality standards and contextual reasoning rather than implying perfectly rational agents—toward pattern-based approximation.

Overall, our findings indicate that delegating evaluations to models transforms how reliability and bias are assessed, replacing human judgment with statistical approximation.

\section*{Results and Discussion}\label{sec2}
To investigate how LLMs perform in practice, we begin by analyzing their classification of online news outlets along the dimensions of reliability and political bias.
Specifically, we evaluate six state-of-the-art models—Deepseek V3, Gemini 1.5 Flash, GPT-4o mini, Llama 3.1 405B, Llama 4 Maverick, and Mistral Large 2—by comparing their outputs to expert human benchmarks from NewsGuard and MBFC. Beyond assessing alignment with expert judgments, we aim to examine the heuristics and decision patterns these models deploy, providing a first approach to investigate the underlying processes that guide model evaluations. To support this analysis, we construct a diverse dataset of 7,715 English-language news outlet domains. The sample spans multiple countries and includes outlets with both national and international reach.

We extract a snapshot of each domain’s homepage, removing nonessential elements (e.g., scripts, styling) to isolate relevant textual content such as headlines and descriptions. This pre-processing step ensures that all LLMs are evaluated on the same textual input a human assessor would plausibly consider.
The final dataset includes 2,286 active domains successfully classified and output in well-formed JSON documents by all models. A detailed breakdown of the data collection and processing is provided in the Methods section.

Beyond analyzing the final labels produced by the models, we examine the strategies they use to generate these judgments, thus providing insight into how LLMs encode and operationalize the notion of reliability. 

We begin our assessment by querying each model using a zero-shot, closed-book approach, providing no examples or explicit definitions of reliability. This setup constrains models to rely on internal representations acquired during training. Our goal is to analyze, through model outputs as a proxy, the heuristics guiding their classifications and to evaluate where their judgments align with or diverge from structured human evaluations.

To move beyond a simple binary classification (Reliable or Unreliable), we prompt each model to assign a political orientation label to each outlet and justify its assessment by generating explanatory keywords. Using a standardized prompt across all six models, this setup enables direct comparison of their outputs and provides insight into how models construct their reliability judgments and how these compare to expert human evaluations.

We also implement an agentic framework in which LLMs autonomously retrieve news outlet pages and follow a structured evaluation pipeline to assess reliability. This approach allows for a controlled comparison between LLMs and human evaluators when given the same task.

\subsection*{LLMs vs. Expert-Driven Assessments}

\begin{figure}[t]
    \centering
    \includegraphics[width=0.6\linewidth]{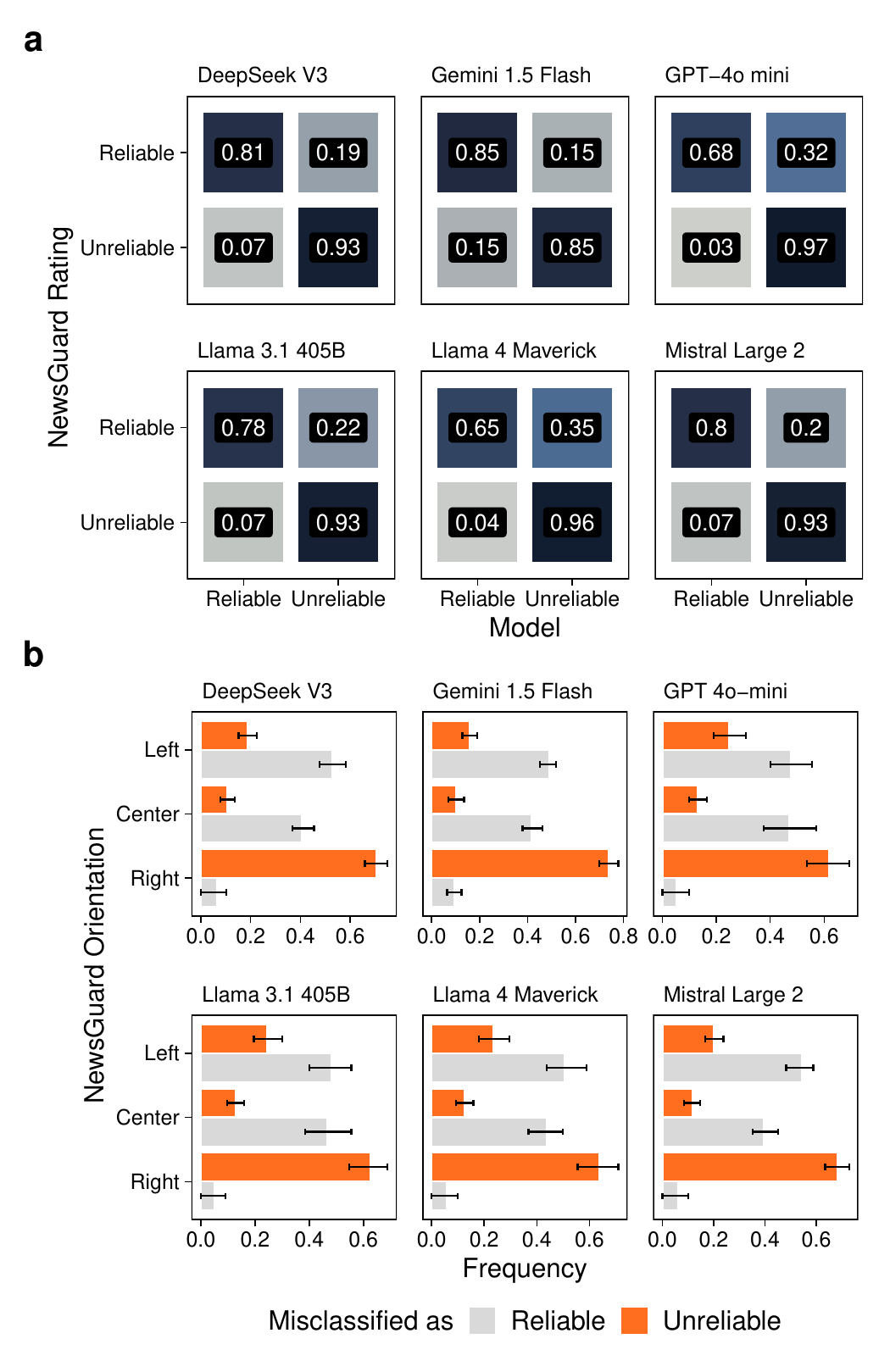}
    \caption{\textbf{LLMs' classification against expert human evaluators.}  (a) Each panel compares how domains rated as “Reliable” or “Unreliable” by NewsGuard are classified by each LLM (Deepseek V3, Gemini 1.5 Flash, GPT-4o mini, Llama 3.1 405B, Llama 4 Maverick, Mistral Large 2). All six models accurately identify Unreliable sources, with agreement ranging from 85 to 97\% across models. However, Reliable domains show greater classification variability, particularly in Llama 4 Maverick and in GPT 4o-mini, which classify a significant portion (35\% and 32\%) as “Unreliable.” (b) We randomly sample 40 domains from each pairing of NewsGuard’s political orientation and reliability rating and compute the average misclassification rate across political orientations over 10,000 resamples. The error bars report the first and third quartiles of the resulting frequencies per group. Compared with NewsGuard, LLMs appear to overestimate or underestimate the reliability of news outlets based on their political orientation. In particular, Right-leaning news outlets tend to be consistently misclassified by the LLMs as unreliable, whereas the Center and Left-leaning as reliable.
    }
    \label{fig:conf_matrix}
\end{figure}

Figure \ref{fig:conf_matrix}a compares the classifications produced by each model with the reliability ratings assigned by NewsGuard. These ratings are not arbitrary but derived from a structured evaluation protocol based on systematic assessments of editorial standards, transparency, and factual accuracy. By contrast, LLMs make decisions without directly accessing these guidelines, relying instead on internal heuristics formed during training.

All six models accurately identify ``Unreliable'' sources, consistently flagging domains that NewsGuard associates with low credibility or lack of transparency. In contrast, classifying ``Reliable'' sources proves more difficult. GPT-4o mini and Llama 4 Maverick, in particular, misclassify 32\% and 35\% of reliable domains, respectively—substantially more than the other models. This asymmetry may reflect the multifaceted nature of NewsGuard’s evaluation criteria (e.g., editorial standards, correction policies, ownership transparency), which may not be fully captured from homepage content alone.
We further assess model alignment with expert judgments by comparing their classifications against the Credibility ratings assigned by Media Bias Fact Check (MBFC) for a subset of 916 overlapping domains.

Consistent with Fig. \ref{fig:conf_matrix}a, LLMs often achieve over 90\% accuracy with MBFC ratings at the extremes: sources labeled Low or High credibility are correctly classified as Unreliable and Reliable, respectively. For Medium credibility sources, however, model performance diverges—both from MBFC and across models. For instance, GPT-4o mini and Llama 4 Maverick classify the majority of these domains as Unreliable (75\% and 77\%, respectively), while Gemini 1.5 Flash produces a more balanced distribution. This pattern is consistent with the interpretation that LLMs rely primarily on clear-cut textual cues when available, while showing lower accuracy on ambiguous or borderline cases. The confusion matrices for each model are provided in Supplementary Figure S1.
Although the models lack explicit access to the evaluation procedures of NewsGuard and MBFC, and are not provided with their methodological criteria, their outputs are broadly consistent with the credibility assessments made by expert human fact-checkers.

We next examine whether misclassifications by LLMs are uniformly distributed across the political orientation labels assigned by NewsGuard, or whether specific orientations are disproportionately affected. To do so, we draw random samples of 40 domains for each combination of NewsGuard’s political orientation and reliability labels—the smallest group size in the dataset—and compute the proportion of reliability misclassifications per group. This sampling procedure is repeated 10,000 times to estimate average misclassification frequencies. 

As shown in Fig.\ref{fig:conf_matrix}b, classification errors are not evenly distributed across the political spectrum. Among domains rated as Reliable by NewsGuard, Right-leaning outlets are consistently misclassified as Unreliable more often than Center or Left-leaning ones, whose reliability tends instead to be overestimated. Importantly, this asymmetry does not indicate that LLMs hold partisan preferences. Rather, consistent with recent work on value alignment and political bias in LLMs (e.g., \cite{motoki2025assessing,hernandes2024llms}), it likely reflects correlations in the training data—for instance, the co-occurrence of extremist rhetoric and misinformation—rather than an explicit ideological stance. As a result, models may overgeneralize, conflating legitimate right-leaning journalism with toxic or conspiratorial sources when linguistic markers overlap.

We also evaluate how the political orientation labels assigned by the models compare to those from human annotators. All six LLMs show strong agreement with NewsGuard ratings, as illustrated in Supplementary Figure S2, with substantial overlap across the political spectrum. Some discrepancies arise from the finer-grained set of labels used by the models compared to NewsGuard’s coarser taxonomy. This alignment is further supported by comparisons with MBFC’s ``Bias Rating'', focusing on strictly political classifications.

\subsection*{Explaining Reliability Ratings with Keywords}

\begin{figure*}[t!]
    \centering
    \includegraphics[width=0.95\linewidth]{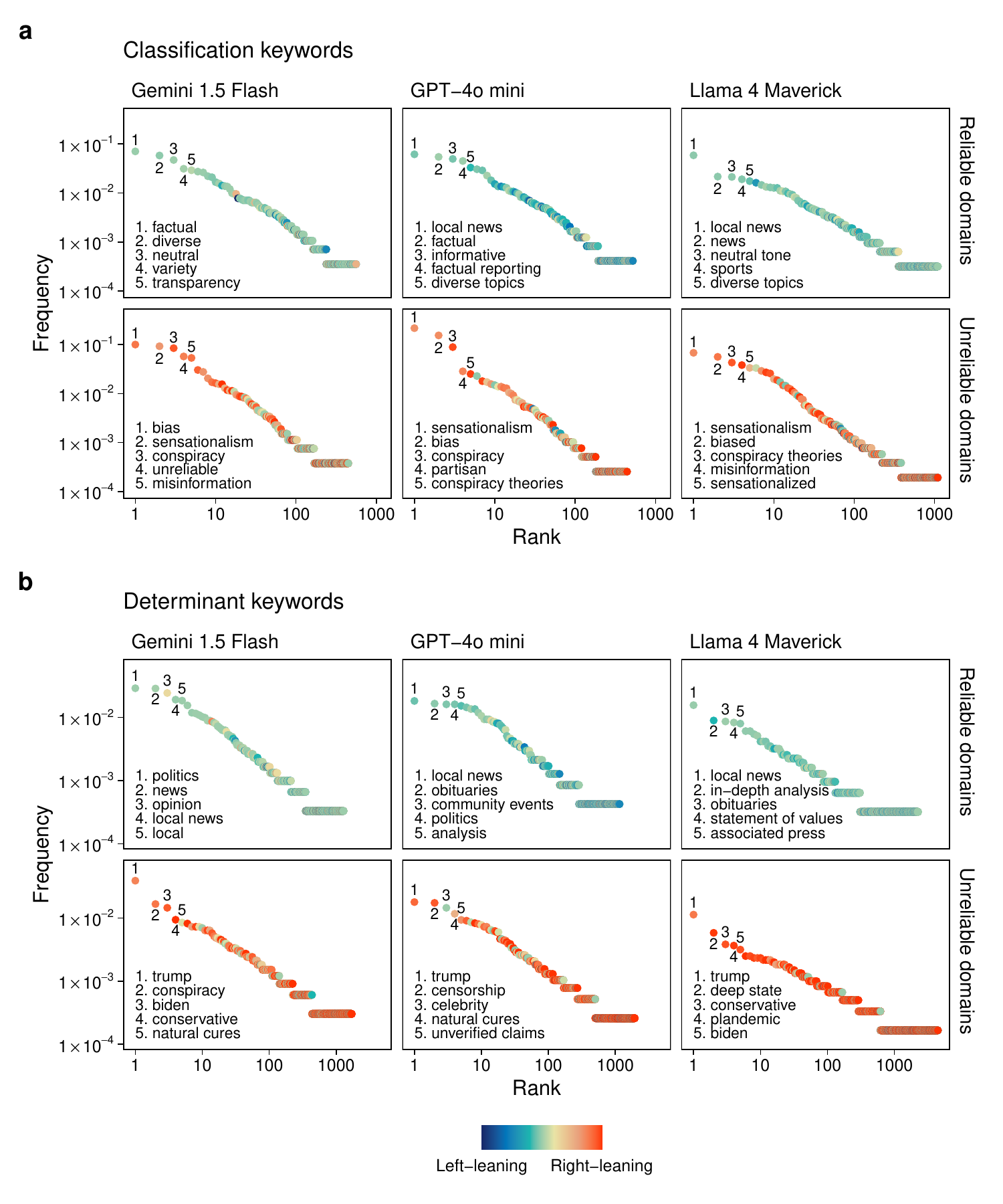}
    \caption{\textbf{Rank-frequency distributions of keywords used by each LLM to describe domains}. Each panel presents the most frequently used classification (a) and determinant (b) keywords for Reliable and Unreliable domains. Only the five most common keywords per panel are labeled to enhance readability. The color gradient represents the inferred political orientation of each keyword, ranging from Left-leaning to Right-leaning, based on the political leaning of the domains they are most frequently associated with. Right-leaning keywords appear almost exclusively in descriptions of Unreliable domains, whereas politically neutral or Left-leaning keywords are more characteristic of Reliable domains. All distributions exhibit heavy-tailed behavior, as indicated by their roughly linear shape on a log–log scale, where a small set of highly frequent keywords dominate the descriptions, while the majority appear less frequently. This indicates that LLMs produce consistent markers when explaining their reliability evaluations.}
    \label{fig:keywords_rank_frequency}
\end{figure*}

We examine the factors driving LLMs' reliability judgments by analyzing three sets of keywords generated by each model for every news outlet, alongside their assigned reliability and political orientation labels. These lexical cues provide indirect evidence about the heuristics models used to approximate credibility in the absence of explicit scoring guidelines. Unlike human evaluators, LLMs rely on implicit heuristics—emerging from patterns in their training data—underscoring the importance of analyzing the associations reflected in their outputs.

For each domain, every LLM generates three distinct sets of keywords: (i) classification keywords, reflecting the rationale behind the assigned reliability rating; (ii) determinant keywords, extracted directly from the domain’s homepage and considered critical for the classification; and (iii) summary keywords, broadly capturing the overall content of the homepage. All keywords are converted to lowercase prior to analysis. We impose no constraint on the number of keywords generated, allowing us to observe each model’s typical lexical output and assess whether keyword volume varies by reliability label or across models. Constraining output length would risk limiting the models’ expressive capacity and reducing the interpretability of their reliability assessments.

To quantify the political leaning of each keyword, we compute the average orientation of the domains in which it appears, using the political labels assigned by the models. These categorical labels are mapped onto a numerical scale from $-1$ (Left) to $1$ (Right), with intermediate values assigned as follows: $-0.5$ (Center-Left), $0$ (Center), and $0.5$ (Center-Right).

To analyze keyword usage, we construct rank-frequency distributions separately for each model, keyword type, and reliability label. In these distributions, terms are ordered by frequency of occurrence, with rank 1 assigned to the most frequent. 

To streamline the presentation, we focus on three representative models—Gemini, GPT, and Llama 4—with results for the remaining models reported in the Supplementary Information.
Figure \ref{fig:keywords_rank_frequency} shows the rank-frequency distributions of classification and determinant keywords by model and reliability label. All models exhibit a heavy-tailed pattern, indicating that a common core set of linguistic markers is consistently associated with the classification procedure. This is consistent with natural language corpora, where few words occur frequently while most appear rarely \cite{di2024patterns}.

As shown in Fig. \ref{fig:keywords_rank_frequency}, classification keywords capture linguistic markers associated with model classifications in reliability assessment. Reliable domains are frequently linked to terms denoting neutrality, transparency, and factual reporting, suggesting a focus on balanced communication and professional presentation. In contrast, unreliable domains are consistently associated with terms such as “misinformation”, “conspiracy”, and “bias”, consistent with patterns often linked to sensationalism and partisanship. These patterns indicate that LLM classifications exhibit structured linguistic heuristics that partially mirror human evaluative criteria.

Determinant keywords offer further insight into the patterns underlying model classifications. Reliable domains are often associated with references to editorial standards and institutional transparency. GPT-4o mini and Llama 4 notably emphasize ``local news'', suggesting that community-based reporting appears as a marker of credibility. In contrast, unreliable domains are consistently linked to politically charged terms, with keywords such as ``trump'', ``biden'' and ``deep state'' recurring prominently, suggesting that politicized content is often associated with unreliability.

Additionally, Fig. \ref{fig:keywords_rank_frequency} highlights an asymmetry in the political connotation of keywords: right-leaning terms are more prevalent in descriptions of unreliable sources, whereas neutral or left-leaning terms appear more often in association with reliable domains.

\begin{figure*}[t!]
    \centering
    \includegraphics[width=0.95\linewidth]{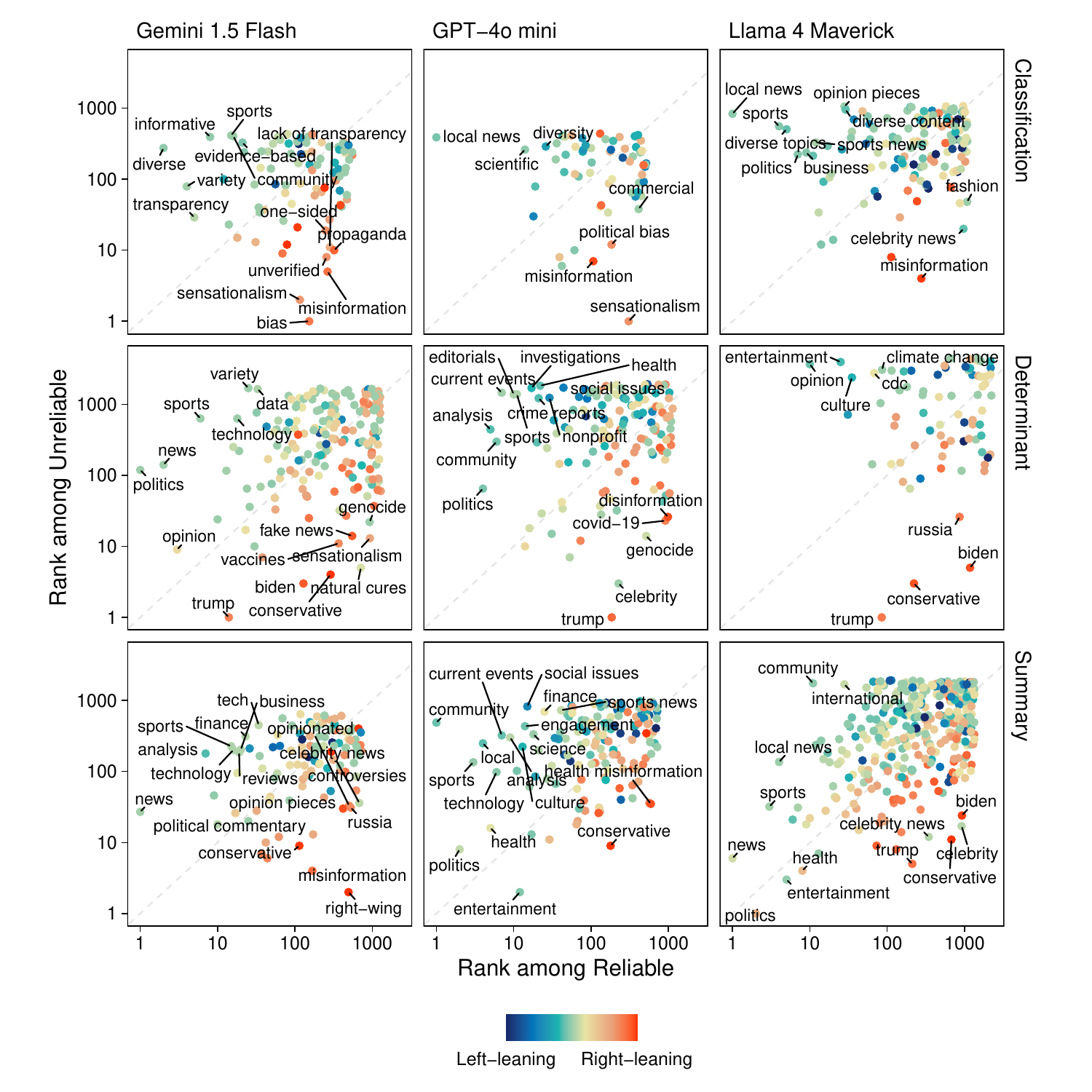}
    \caption{\textbf{Keywords’ rank among Reliable and Unreliable domains.} We label only keywords sufficiently distant from the diagonal, meaning they are predominantly used to describe reliable or unreliable domains rather than being evenly distributed across both classifications. Additionally, we label the top 5 keywords per reliability rating. The color gradient represents the inferred political orientation of each keyword, from Left-leaning to Right-leaning, based on the domains with which they are most frequently associated. While summary keywords (Bottom row) appear with similar frequency in both reliable and unreliable domains, classification and determinant keywords (Top and Middle rows) exhibit sharper separation. This result suggests that reliable and unreliable sources may cover similar topics but differ in framing tone or contextual emphasis. Notably, keywords related to transparency, objectivity, and credibility are more common among reliable domains. At the same time, sensationalist and politicized terms such as “misinformation,” “propaganda,” and “bias” are frequently linked to unreliable sources.
    }
    \label{fig:keywords_rank_rel_v_unrel}
\end{figure*}

Keywords used to describe both reliable and unreliable domains are shown in Fig. \ref{fig:keywords_rank_rel_v_unrel}, which compares their rank across the two classifications. The farther a keyword lies from the diagonal, the more distinctive it is of either reliable or unreliable domains.

When examining classification and determinant keywords, clear differences emerge between the two reliability groups. Reliable domains are associated with terms such as ``local news'', ``scientific'', ``diverse'', and ``evidence-based'', while unreliable classifications involve more controversial or politically charged terms, including politician names (e.g., ``trump'', ``biden'') and topics such as ``genocide'' and ''vaccines''.
In contrast, summary keywords—describing the overall content of a domain—show substantial overlap between the two groups. This suggests that reliable and unreliable sources often cover similar topics, and that the distinction lies less in what is covered than in how it is presented.  We also find that terms with no clear semantic polarity tend to be unevenly distributed across classes, suggesting that model judgments may be associated with framing and usage context rather than meaning alone.

\subsection*{Human and LLM Credibility Assessment}

\begin{figure}[t!]
    \centering
    \includegraphics[width=0.8\linewidth]{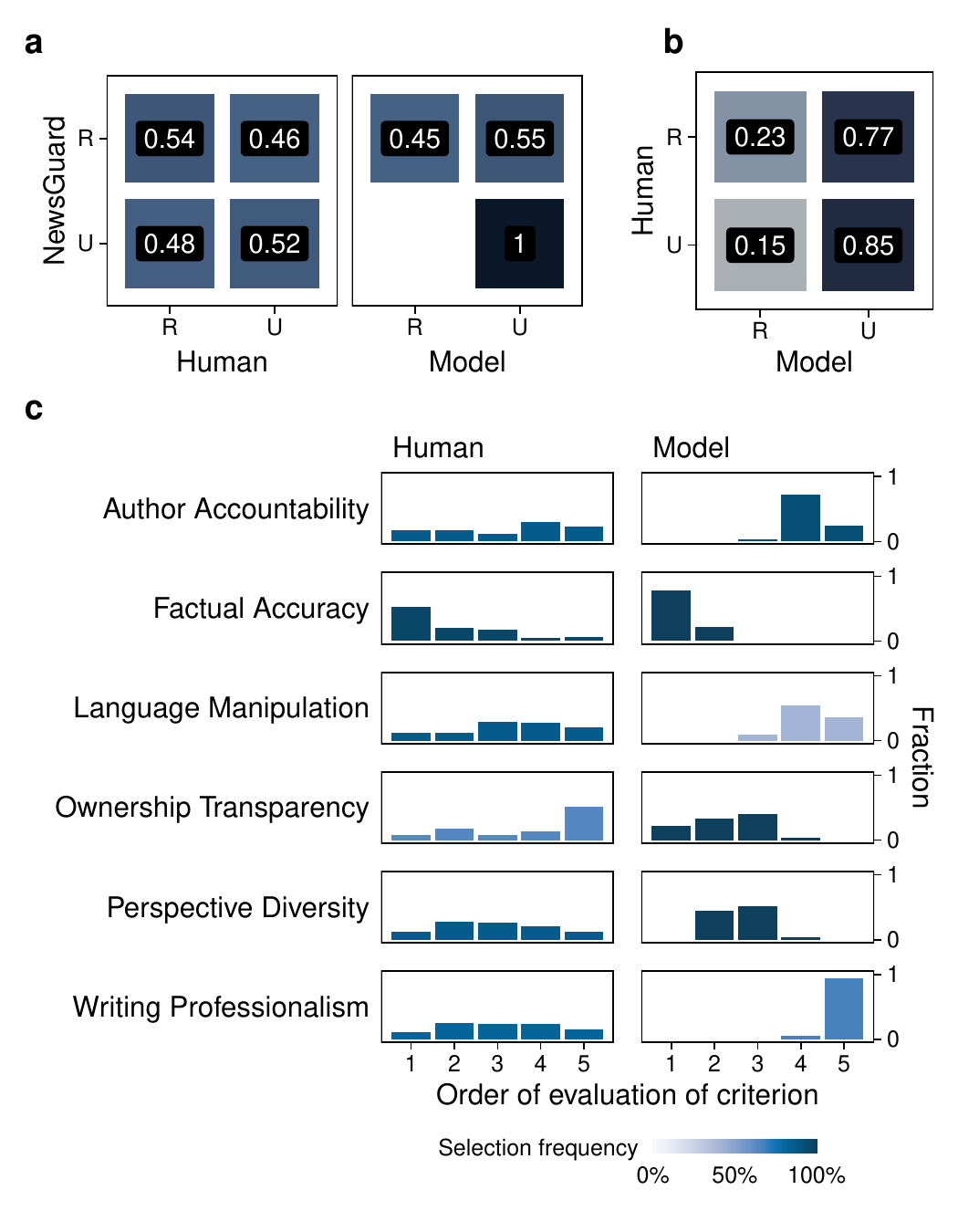}
    \caption{\textbf{Reliability evaluations by Gemini-powered LLM agents and non-expert humans in a controlled experimental setting.} (a) The two panels compare humans’ and agents’ reliability ratings against NewsGuard’s classifications. Models consistently identify all Unreliable (U) sources and struggle with the Reliable (R). In contrast, humans show little to no alignment with NewsGuard, for both reliable and unreliable domains. (b) Confusion matrix of ratings provided by humans and agents, with the human ratings used as the ground truth. The two show strong agreement on unreliable sources, while 77\% of sources rated as reliable by humans are considered unreliable by the LLM. (C) Distributions of order choices for each criterion by humans (Left) and models (Right). The human distributions appear more uniform than those of the models, indicating that most criteria are roughly equally likely to appear in any position compared to LLMs.}
    \label{fig:human_v_model}
\end{figure}

Our previous analysis shows that LLMs often achieve high agreement with expert evaluations from NewsGuard and MBFC. This suggests that, despite lacking access to structured evaluation criteria, their outputs reflect heuristics that approximate human judgments. Yet a key question remains: what procedures or proxies underlie these evaluations?

A critical observation emerges when models are prompted with only a domain URL, without access to any homepage content \cite{yang2023large}. We find that, even under these minimal conditions, LLMs generate reliability ratings that broadly align with expert assessments. For instance, Gemini achieves an F1-score of 0.78—only slightly below the 0.86 obtained with full HTML input—while GPT reaches 0.77, compared to 0.79. 

This raises a foundational question: are LLMs engaging in content-specific evaluation or relying primarily on statistical associations learned during training? If a model can classify a domain without analyzing its content, it becomes difficult to disentangle content-based assessment from statistical recall. More broadly, such behavior is consistent with the interpretation that model assessments may be shaped by prior knowledge about the news outlet rather than by content-specific evaluation.

To analyze this issue, we introduce a structured agentic workflow that enables a direct comparison between LLMs and human evaluators. Rather than treating models as black boxes producing binary outputs, we instantiate a multi-agent system designed to simulate the procedural steps involved in human evaluation—retrieving, processing, and integrating information before rendering a judgment. An agentic pipeline is well-suited to this task, as it provides LLMs with access to modular tools (i.e., external functions implemented in code) and enables the composition of deterministic multi-step workflows, where intermediate outputs from one LLM can be passed to the next. To implement this workflow effectively, we focus on Gemini 2.0 Flash, which optimally supports the tooling required for implementing the agentic pipeline. Given the consistency observed across models in prior sections, we expect the insights to generalize.

The protocol, detailed in the Methods section, unfolds as follows. First, an LLM agent selects five out of six predefined evaluation criteria and ranks them by importance. This selection happens before the LLM is exposed to any content from the news outlet apart from its URL, which is necessary to start the pipeline. However, further experiments we conducted, presented in Supplementary Information, suggest that exposure to the URL does not appear to influence this selection. Additionally, the criteria are provided to the model in randomized order. Two additional agents then retrieve content: one downloads the homepage, while the other extracts up to two articles deemed informative for assessing credibility, using the URLs of articles found in the homepage. This content is collected using a tool available to agents that downloads webpages in a well-structured Markdown document. The number of retrieved articles is not fixed, allowing the agent to autonomously select zero, one, or two. Subsequently, five additional agents independently evaluate the selected criteria, each using only the materials retrieved in the prior step. Their outputs include both a numeric reliability rating from 1 to 5 and a written assessment. A final agent then aggregates these assessments into a structured summary and produces a binary reliability classification, based upon the other agents' evaluations. This workflow enables a direct procedural comparison between LLMs and humans, both following the same evaluation steps in a controlled setting.

To facilitate this comparison, we replicated the agentic protocol in a human subject experiment. We recruited $N=50$ participants and tasked them with assessing the reliability of online news outlets using the same structured procedure: selecting five criteria, ranking them, browsing the homepage and up to two articles, and finally producing a binary reliability judgment. Full details on participant recruitment and setup are provided in Materials and Methods.

Although a total of 37 Italian-language news outlet domains were evaluated by the recruited human participants, only 27 of these could also be evaluated using our agentic workflow, as reported in Supplementary Table S2. This limitation was due to webpage retrieval requests being blocked by some domains. Figure \ref{fig:human_v_model} summarizes the results for this subset of domains, while in Supplementary Fig. S7 we report the results from human evaluators for the full set. Panel (a) compares the reliability ratings produced by LLMs and humans against NewsGuard’s classifications. LLMs, even when constrained to a more rigid evaluation pipeline and operating on non-English content, maintain consistency with earlier findings, though with lower accuracy. In contrast, human participants show no meaningful alignment with NewsGuard: reliable and unreliable domains are classified with roughly equal probability, suggesting that non-expert evaluators rely on different and less consistent indicators. Panel (b) compares LLM and human ratings directly, treating human judgments as a referential baseline. While both groups agree on which outlets are unreliable, major divergence arises for reliable sources: LLMs classify as unreliable almost 80\% of the domains rated as reliable by humans.

This asymmetry aligns with prior evidence that individuals are often more prone to reject accurate information than to believe falsehoods. A recent meta-analysis covering over 195,000 participants across 67 studies confirms this tendency—commonly referred to as skepticism bias—and shows that non-experts are especially likely to misclassify true information as false \cite{pfander2025spotting}. Our findings suggest that this bias persists even under structured and controlled evaluation.
By directly integrating our experiment with this literature, we show that LLMs reproduce some behavioral regularities observed in humans (e.g., skepticism bias) while diverging sharply in the indicators that guide their judgments.

To explore the basis of this divergence, we examine the evaluation criteria selected by each group. Figure \ref{fig:human_v_model}c shows the ranking distributions across the six available criteria (Table \ref{Table 1}). Both groups consistently prioritize ``Factual Accuracy'', defined as the extent to which the website’s content is ``accurate and free from false or misleading information''. Nearly all participants—human and LLM—selected this criterion and ranked it first. However, this convergence masks deeper differences in reasoning processes. For human participants, prioritizing accuracy likely reflects deliberative reasoning based on content comprehension and analytical judgment \cite{pennycook2019lazy}. In contrast, for LLMs this emphasis appears to be operationalized via lexical associations and patterns learned during training, as suggested by proxy analyses.

Beyond this shared top criterion, the hierarchies diverge sharply. LLMs consistently rank ``Ownership
Transparency'' among the top three criteria, while humans rarely select it and often place it last. This is consistent with findings that most individuals—unlike professional fact-checkers—rarely engage in “lateral reading” to verify sources or consult external indicators of trustworthiness \cite{wineburg2019lateral}. LLM outputs display patterns consistent with more structured verification behavior, possibly reflecting exposure to factuality benchmarks during training. Human evaluators, by contrast, often rely on surface-level cues and intuitive judgments—especially when cognitive resources are limited\cite{fiske2018social,tversky1974judgment}.

Compared to LLMs, human participants tend to prioritize rhetorical and stylistic cues, such as ``Language Manipulation'' and ``Writing Professionalism''. Emotional tone is often interpreted as a sign of persuasive intent rather than factual reliability \cite{metzger2010social}, and fluency of expression is linked to perceived truthfulness through the processing fluency heuristic \cite{reber1999effects} and the well-documented illusory truth effect, the tendency to perceive repeated information as more likely to be true regardless of its actual accuracy \cite{fazio2015knowledge}. Given that LLMs lack affective or metacognitive grounding, these features appear to carry little epistemic weight in their evaluations, functioning mainly as superficial markers.
Taken together, this integration with prior literature suggests that LLMs approximate some dimensions of human evaluation while omitting others.
Overall, this agentic comparison enables a more granular understanding of how LLMs appear to operationalize credibility. What emerges is not just a difference in outcomes but a difference in the observable patterns through which evaluation processes are instantiated. 

Humans blend intuitive and analytical processes \cite{kahneman2002maps}; LLM outputs, by contrast, reflect different patterns of evaluation shaped by statistical learning. Recognizing this distinction does not diminish the utility of LLMs in evaluative tasks, but it clarifies the boundaries of their competence. Our findings point to a shift from content-based deliberation toward plausibility-driven approximations when judgment is delegated to automated systems—a dynamic we term epistemia, where statistical plausibility risks replacing deliberative reasoning with the illusion of knowledge rather than its verification. Future research could link behavioral experiments, cognitive models, and automated evaluation pipelines to map where human and machine heuristics converge and where they fundamentally diverge.

\section*{Conclusions}\label{sec12}
This study examined how Large Language Models (LLMs) operationalize core evaluative concepts—such as reliability and bias—when tasked with assessing online news outlets, comparing their judgments to expert benchmarks and to human participants following the same protocol. While model outputs often align with expert classifications—especially when flagging unreliable sources—this apparent agreement conceals a deeper divergence in the evaluative mechanisms themselves, raising broader questions about delegating judgment to automated systems in information environments already shaped by infodemics and platform-driven filtering.

Consistent with our findings, LLMs operate through lexical associations, statistical priors, and structural cues, rather than genuine contextual interpretation. This statistical approximation produces systematic asymmetries: right-leaning outlets are disproportionately classified as unreliable. When comparing LLM outputs to human judgments, we find further divergence: models display a pattern reminiscent of the skepticism bias observed in humans—an over-rejection of accurate information documented in large-scale studies \cite{taber2006motivated}.

By integrating our human experiment with existing work on credibility assessment and cognitive biases \cite{metzger2010social, rieh2010credibility, pfander2025spotting}, we show that LLMs partially replicate behavioral regularities identified in psychology while relying on fundamentally different evaluative mechanisms. This link clarifies that what emerges is not simply an accuracy gap, but a structural shift in how evaluation itself is operationalized when judgment is delegated to automated systems.
Both groups prioritize ``Factual Accuracy'' as the most important criterion, but the underlying mechanisms differ. Human participants likely interpret accuracy through content comprehension and pragmatic reasoning; LLMs, instead, derive it from statistical regularities encoded during training. This distinction becomes clearer in the secondary criteria: LLMs emphasize ``Ownership Transparency''—aligning with professional fact-checking protocols—while humans give more weight to stylistic and rhetorical features such as tone and writing fluency. These preferences mirror known heuristics like processing fluency, whereby clarity and emotional neutrality enhance perceived truth.

Such discrepancies underscore a structural difference between intuitive, context-aware human evaluation and the pattern-based, procedural mechanisms of LLMs. As these systems are increasingly embedded in decision-making pipelines—moderation, classification, prioritization—it becomes critical to assess not just whether their outputs appear reasonable, but how their internal procedures operationalize normative categories like reliability and bias.
This is especially urgent in an information ecosystem already marked by infodemics, where the oversupply of low-quality or contradictory information erodes trust and amplifies polarization \cite{del2016spreading}. In this context, the rise of what we term epistemia—the illusion of knowledge emerging when plausibility replaces verification—illustrates the risk that statistical approximation could displace deliberative reasoning if adopted uncritically.

Whereas the first wave of research on social media emphasized the volume and velocity of information flows \cite{lazer2018science,cinelli2020covid}, our findings highlight a second phase in which the problem shifts from information overload to the nature of judgment itself, as automated pipelines introduce epistemic opacity into evaluative processes.
Our structured agentic framework enables such comparison by aligning inputs, tasks, and justification protocols across humans and LLMs. While our sample size limits generalizability, the controlled design and multilingual robustness provide a solid foundation for further research. Future work should examine how this epistemic shift interacts with governance, transparency, and human oversight, especially as automated judgment pipelines expand beyond content moderation into law, policy, and scientific evaluation.

In sum, the apparent alignment between LLMs and expert judgments may mask only superficial output convergence. Delegating evaluative tasks to these systems risks embedding frameworks driven by lexical and statistical associations rather than deliberative reasoning, amplifying existing information pathologies. Addressing this shift requires transparency, human oversight, and potentially new training paradigms that explicitly disentangle factual reliability from ideological or stylistic cues. Hybrid approaches that combine statistical models with explicit reasoning criteria or retrieval-based evidence may offer a promising way forward.

\section*{Materials and Methods}\label{sec:methods}
\subsection*{Data collection and pre-processing}\label{sec:methods_data}
All data was collected by downloading the HTML homepages of domains rated by NewsGuard as ``reliable'' or ``unreliable'', using the \texttt{requests} library available on Python \cite{githubrequests}. These domains have been selected among outlets reported by NewsGuard as English-speaking, based in an English-speaking country (US, GB, CA, AU, NZ), and with a National or International focus. Not all domains could be downloaded, as many were either no longer active at the time of downloading, only accessible from specific regions, or designed in such a way as to render automatic scraping difficult.

The downloaded pages are then filtered to retain only the information relevant to the LLMs to evaluate the reliability of each domain. This process involves removing unnecessary HTML components, such as scripts, styles, and navigation elements, using the \texttt{beautifulsoup4} library on Python \cite{crummyBeautifulSoup}. The resulting document almost exclusively includes textual content, particularly related to news or information about the outlet. While domain-specific pre-processing techniques can also be implemented, we opted for a more conservative, universal approach that can easily be applied to any news outlet's homepage. 
The procedure yields a dataset of 2,302 domains with at least 100 characters of content. Finally, we retain only the 2,286 domains successfully classified and output in well-formed JSON by all six LLMs, with 1,188 rated by NewsGuard as ``reliable'' and the remaining 1,098 as ``unreliable''.

\subsection*{Model prompting}\label{sec:methods_prompt}

\begin{figure}[t]
\scalebox{0.9}{
    \begin{tcolorbox}[colback=gray!10!white, colframe=gray!70!black, sharp corners=south, boxrule=0.8pt, width=1\textwidth]
    Given the HTML of the homepage of a news outlet, you classify it as either `reliable' or `unreliable' based on its content. Focus on the entire homepage, rather than just the main content or story that is presented. Make your judgment solely on the contents of the webpage.
    
    \vspace{0.5em}
    
    What rating (`reliable', `unreliable') would you assign to the news outlet with the homepage in the following HTML? I want the output as a JSON with the following attributes:
    \begin{itemize}
        \item `rating': your overall classification rating (`reliable', `unreliable')
        \item `text': your raw response
        \item `orientation': political orientation of the contents of the webpage (`left', `center-left', `center', `center-right', `right')
        \item `classification\_keywords': a list of keywords to justify your reasoning for classification
        \item `determinant\_keywords': a list of keywords from the webpage that were determinant for your classification
        \item `summary\_keywords': a list of keywords that summarize the contents of the webpage
    \end{itemize}
    
    \vspace{0.5em}
    
    HTML:
    
    \texttt{\{scraped HTML\}}
    \end{tcolorbox}
    }
    \caption{Prompt used for all LLMs when provided the scraped HTML homepage.}
    \label{fig:prompt_html}
\end{figure}

We follow a zero-shot, closed-book approach to prompting for all six LLMs we experiment with, by providing no example classifications and no definitions of ``reliability'', respectively. Additionally, for all models, we use the system prompt \textit{``You are a service that gives a reliability rating to news outlets.''} and set the temperature parameter to 0. Our prompt, which we use for all models, is split into two parts. 

The first part gives the model basic instructions regarding the classification task, providing no context to the concept of ``reliability'' beyond mentioning the terms ``reliable'' and ``unreliable''. In the second part, we provide the models with specific instructions concerning their output, specifying both the required content and its structure. Figure \ref{fig:prompt_html} reports the exact prompt used for all models.

Gemini 1.5 Flash, DeepSeek V3, and GPT-4o mini were prompted by performing calls to the official API endpoints made available by Google, DeepSeek, and OpenAI, respectively. Llama 3.1 405B was prompted through SambaNova Cloud, while Llama 4 Maverick and Mistral Large 2 were accessed via OpenRouter, both third-party platforms.

Queries sent to DeepSeek, GPT, Llama 3.1, and Mistral were truncated to ensure they fit within the models' context length (128,000 tokens for all), which is the maximum number of tokens they can process at once. Specifically, the scraped webpages provided to these models were limited to the first 50,000 characters. However, this truncation affected less than 2\% of the domains.

Each domain was evaluated individually, as simultaneous classification of multiple inputs may introduce unwanted bias. For example, reliability might be assessed relative to the specific subset of domains provided in the query, rather than based on the model's inherent notion of ``reliability''. 

If the output political orientation label fell outside the specified 5-point scale, we reassigned it to ``Center''. For all models, this affected none or fewer than 1\% of the domains, except for Mistral Large 2, where approximately 4\% of the domains received an out-of-scale label.

When evaluating the LLMs' ability to classify news outlets using only their domain names, we slightly altered the prompt in Fig. \ref{fig:prompt_html} by substituting the first paragraph with the text ``\textit{Given the domain of a news outlet, you classify it as either `reliable' or `unreliable' based on its content.}'', and by replacing all other occurrences of `HTML' with `URL'.

\subsection*{Agentic workflow}
\label{sec:methods_agents}
We implemented the agentic workflow for outlet reliability classification with Google's Agent Development Kit (ADK) \cite{githubADK}. ADK is a Python toolkit that allows for developing and orchestrating agentic systems. In particular, for our study, we developed a multi-agent system where each sub-task in the reliability evaluation procedure is delegated to a dedicated agent. The implementation firstly relies on what ADK calls a ``workflow agent'', specifically a ``Sequential Agent''. This is not an LLM-powered agent, but rather a component in ADK that allows for orchestrating the other agents in a deterministic manner. Specific to our case, this agent is what enables our procedure to follow a well-defined path, as agents are called one after the other. The initial prompt used to start the workflow is \textit{``Select the most appropriate reliability criteria among those provided and evaluate the reliability of \{URL\}''}. We provide all additional prompts used in this workflow in the Supplementary Information.

The first agent is a ``code agent'', that is, an LLM agent that performs actions by writing code and calling tools written in a programming language \cite{wang2024code}, Python in our case. Specifically, it is a tool-augmented agent that is tasked with selecting five criteria to evaluate from a list of six, and to rank them from \emph{``most to least important for assessing reliability''}. To avoid our chosen order of criteria influencing the model's decision, we do not include the list directly in the prompt. Instead, the agent calls a Python function that returns the criteria in a randomized order.

The next two agents, which are also code agents, are responsible for retrieving all data required for the reliability evaluation. The first one is tasked with retrieving the domain's homepage as a structured Markdown document, by leveraging the \texttt{markitdown} Python library \cite{githubMarkitdown}. Contrary to the data collection process used for our first analyses, in this case, we are not just interested in the textual content of the page, but also in the URLs of the news articles presented on the homepage. The Markdown format is particularly effective at concisely structuring this information in a way that can be easily understood by LLMs. This approach enables the second code agent to analyze the scraped homepage and assess whether to collect up to two articles by scraping their corresponding webpages. All this content is thus stored in the workflow's state to allow all subsequent agents to read it.

Once the criteria are selected and the data collected, the workflow activates all agents tasked with evaluating the criteria, assigning one criterion per agent. These LLM agents are not provided with any coding tool, meaning they function analogously to any other LLM assistant when given a prompt. Further, they produce their assessment independently from each other. Each agent is asked to output a rating from 1 to 5, with higher scores corresponding to higher reliability, a written summary explanation with examples and quotes from the analyzed content, and a binary flag indicating whether the agent also analyzed the downloaded articles for evaluating the criterion. Additionally, we instruct a separate agent to analyze the news outlet's political orientation on a 5-point scale from Left to Right. 

The information produced by all agents is then reviewed by a final LLM agent, which is tasked with assigning an overall binary reliability rating: ``reliable'' or ``unreliable''.

We set the temperature to 0 for all LLM agents, except for the one responsible for selecting the evaluation criteria. For this agent, we instead used a temperature of 1 to introduce more variability in the outputs. However, as shown in Supplementary Information Figures S5 and S6, the prioritization of criteria observed in Fig. \ref{fig:human_v_model}c remains largely consistent across the full range of temperatures available for Gemini 2.0 Flash, likely because the task is highly constrained rather than open-ended.

\subsection*{Experimental design}
\label{sec:methods_human_experiment}
Here we provide details about the experimental setting with human participants.

\subsubsection*{Participants}
A total of 50 participants (28 females, 22 males; $\mu_{\text{age}} = 28.4$, $\sigma_{\text{age}} = 9.6$) took part in the in-person experiment sessions at the Department of Psychology of Sapienza University of Rome. All participants were recruited online through advertisements distributed via social media platforms (e.g., LinkedIn, Facebook) and by snowball sampling. 
Eligibility was limited to native Italian-speaking adults (aged $\geq18$) with normal or corrected vision. All human participants were non-experts with no prior training in credibility assessment, ensuring that judgments reflected naïve, uncoached heuristics rather than professional fact-checking protocols.

\subsubsection*{Materials and Procedure}

\par
\par
  \begin{table}[t]
    \centering
    \scalebox{0.95}{
    \begin{tabular}{|>{\raggedright\arraybackslash}p{0.3\linewidth}|>{\raggedright\arraybackslash}p{0.7\linewidth}|}
        \hline
                Criterion & Question
\\ \hline
        Author Accountability & To what extent does the site provide the names of content authors, along with their biographies or contact information?
\\ \hline
        Factual Accuracy& To what extent do you believe the content presented on the site is accurate and free from false or misleading information?
\\ \hline
        Language Manipulation& To what extent does the site use emotionally charged, exaggerated, or manipulative language?
\\ \hline
        Ownership Transparency& To what extent does the site clearly declare who owns it and who provides funding for it?
\\ \hline
        Perspective Diversity& To what extent does the site present content offering diverse perspectives without ideological or political bias?
\\ \hline
        Writing Professionalism& To what extent does the site adhere to grammatical rules and use a clear, consistent, and professional writing style?
\\\hline
    \end{tabular}
    }
    \caption{Criteria and corresponding questions}
    \label{Table 1}
\end{table}
The procedure was carried out in a controlled laboratory setting and administered to participants using a Lenovo laptop with a 15.6-inch screen. The experiment was conducted through the Google Chrome web browser to ensure compatibility and correct display of the stimuli. Before taking part in the testing phase, participants provided informed consent. 

The testing phase consisted of two parts: a criteria selection task and an evaluation task.

In the first part, participants were given a set of six criteria in question form (see Table \ref{Table 1}) for assessing the reliability of news domains. 

Participants were asked to identify five criteria they deemed the most relevant for evaluating news credibility and to rank them from most (1) to least (5) relevant. They were asked to exclude one criterion that they deemed non-relevant for the assessment, in order to encourage critical prioritization and to avoid uniform ratings across all criteria. 
In the second part, participants were shown six real and publicly accessible Italian-language news outlets, one at a time. The order in which the websites were presented was randomized across participants. Each participant was instructed to freely navigate the websites and to read up to two full articles to deepen their evaluation. Subsequently, they responded to each question corresponding to the criteria selected in the first phase on a 5-point Likert scale (1 = not at all, 5 = completely) for each website. Additionally, participants provided a binary judgment (Yes/No) concerning the overall reliability of the website, answering the question \textit{``Is this website reliable?''}.
A time limit was imposed for the evaluation of each website to standardize the procedure across participants. The full testing session lasted approximately 15 to 20 minutes. Instructions and interface were given in Italian.
All stimuli were tested before the experimental procedure to ensure usability.

\bmhead{Acknowledgements}
We really thank Luciano Floridi, Mattia Samory, Giovanni Pezzulo and the Hypnotoad for precious insights during the development of the work. The work is supported by IRIS Infodemic Coalition (UK government, grant no. SCH-00001-3391), SERICS (PE00000014) under the NRRP MUR program funded by the European Union - NextGenerationEU, project CRESP from the Italian Ministry of Health under the program CCM 2022, PON project “Ricerca e Innovazione” 2014-2020, and PRIN. 
This work was supported by the PRIN 2022 “MUSMA” - CUP G53D23002930006 - Funded by EU - Next-Generation EU – M4 C2 I1.1.

\bmhead{Ethics Statement}
The study protocol was approved by the Ethics Committee for Transdisciplinary Research of Sapienza University of Rome (Protocol ID: 339/2025).

\bmhead{Author contribution}
W.Q. designed research; E.L., J.N., and N.D.M. performed data extraction and analysis; A.S. and R.A. designed and conducted the human experiments; M.C., V.C., C.R.-A., and W.Q. supervised the work; and all authors contributed to writing the manuscript and provided critical feedback.



\begin{thebibliography}{61}
\ifx \bisbn   \undefined \def \bisbn  #1{ISBN #1}\fi
\ifx \binits  \undefined \def \binits#1{#1}\fi
\ifx \bauthor  \undefined \def \bauthor#1{#1}\fi
\ifx \batitle  \undefined \def \batitle#1{#1}\fi
\ifx \bjtitle  \undefined \def \bjtitle#1{#1}\fi
\ifx \bvolume  \undefined \def \bvolume#1{\textbf{#1}}\fi
\ifx \byear  \undefined \def \byear#1{#1}\fi
\ifx \bissue  \undefined \def \bissue#1{#1}\fi
\ifx \bfpage  \undefined \def \bfpage#1{#1}\fi
\ifx \blpage  \undefined \def \blpage #1{#1}\fi
\ifx \burl  \undefined \def \burl#1{\textsf{#1}}\fi
\ifx \doiurl  \undefined \def \doiurl#1{\url{https://doi.org/#1}}\fi
\ifx \betal  \undefined \def \betal{\textit{et al.}}\fi
\ifx \binstitute  \undefined \def \binstitute#1{#1}\fi
\ifx \binstitutionaled  \undefined \def \binstitutionaled#1{#1}\fi
\ifx \bctitle  \undefined \def \bctitle#1{#1}\fi
\ifx \beditor  \undefined \def \beditor#1{#1}\fi
\ifx \bpublisher  \undefined \def \bpublisher#1{#1}\fi
\ifx \bbtitle  \undefined \def \bbtitle#1{#1}\fi
\ifx \bedition  \undefined \def \bedition#1{#1}\fi
\ifx \bseriesno  \undefined \def \bseriesno#1{#1}\fi
\ifx \blocation  \undefined \def \blocation#1{#1}\fi
\ifx \bsertitle  \undefined \def \bsertitle#1{#1}\fi
\ifx \bsnm \undefined \def \bsnm#1{#1}\fi
\ifx \bsuffix \undefined \def \bsuffix#1{#1}\fi
\ifx \bparticle \undefined \def \bparticle#1{#1}\fi
\ifx \barticle \undefined \def \barticle#1{#1}\fi
\bibcommenthead
\ifx \bconfdate \undefined \def \bconfdate #1{#1}\fi
\ifx \botherref \undefined \def \botherref #1{#1}\fi
\ifx \url \undefined \def \url#1{\textsf{#1}}\fi
\ifx \bchapter \undefined \def \bchapter#1{#1}\fi
\ifx \bbook \undefined \def \bbook#1{#1}\fi
\ifx \bcomment \undefined \def \bcomment#1{#1}\fi
\ifx \oauthor \undefined \def \oauthor#1{#1}\fi
\ifx \citeauthoryear \undefined \def \citeauthoryear#1{#1}\fi
\ifx \endbibitem  \undefined \def \endbibitem {}\fi
\ifx \bconflocation  \undefined \def \bconflocation#1{#1}\fi
\ifx \arxivurl  \undefined \def \arxivurl#1{\textsf{#1}}\fi
\csname PreBibitemsHook\endcsname

\bibitem[\protect\citeauthoryear{Binz et~al.}{2025}]{binz2025should}
\begin{barticle}
\bauthor{\bsnm{Binz}, \binits{M.}},
\bauthor{\bsnm{Alaniz}, \binits{S.}},
\bauthor{\bsnm{Roskies}, \binits{A.}},
\bauthor{\bsnm{Aczel}, \binits{B.}},
\bauthor{\bsnm{Bergstrom}, \binits{C.T.}},
\bauthor{\bsnm{Allen}, \binits{C.}},
\bauthor{\bsnm{Schad}, \binits{D.}},
\bauthor{\bsnm{Wulff}, \binits{D.}},
\bauthor{\bsnm{West}, \binits{J.D.}},
\bauthor{\bsnm{Zhang}, \binits{Q.}}, \betal:
\batitle{How should the advancement of large language models affect the practice of science?}
\bjtitle{Proceedings of the National Academy of Sciences}
\bvolume{122}(\bissue{5}),
\bfpage{2401227121}
(\byear{2025})
\end{barticle}
\endbibitem

\bibitem[\protect\citeauthoryear{\"{O}zer and Perc}{2024}]{zer2024_2}
\begin{barticle}
\bauthor{\bsnm{\"{O}zer}, \binits{M.}},
\bauthor{\bsnm{Perc}, \binits{M.}}:
\batitle{Human complementation must aid automation to mitigate unemployment effects due to ai technologies in the labor market}.
\bjtitle{Istanbul Bilgi University}
(\byear{2024})
\doiurl{10.47613/reflektif.2024.176}
\end{barticle}
\endbibitem

\bibitem[\protect\citeauthoryear{Gao et~al.}{2025}]{gao2025take}
\begin{barticle}
\bauthor{\bsnm{Gao}, \binits{Y.}},
\bauthor{\bsnm{Lee}, \binits{D.}},
\bauthor{\bsnm{Burtch}, \binits{G.}},
\bauthor{\bsnm{Fazelpour}, \binits{S.}}:
\batitle{Take caution in using llms as human surrogates}.
\bjtitle{Proceedings of the National Academy of Sciences}
\bvolume{122}(\bissue{24}),
\bfpage{2501660122}
(\byear{2025})
\end{barticle}
\endbibitem

\bibitem[\protect\citeauthoryear{Yax et~al.}{2024}]{yax2024studying}
\begin{barticle}
\bauthor{\bsnm{Yax}, \binits{N.}},
\bauthor{\bsnm{Anll{\'o}}, \binits{H.}},
\bauthor{\bsnm{Palminteri}, \binits{S.}}:
\batitle{Studying and improving reasoning in humans and machines}.
\bjtitle{Communications Psychology}
\bvolume{2}(\bissue{1}),
\bfpage{51}
(\byear{2024})
\end{barticle}
\endbibitem

\bibitem[\protect\citeauthoryear{Wang et~al.}{2024}]{wang2024survey}
\begin{barticle}
\bauthor{\bsnm{Wang}, \binits{L.}},
\bauthor{\bsnm{Ma}, \binits{C.}},
\bauthor{\bsnm{Feng}, \binits{X.}},
\bauthor{\bsnm{Zhang}, \binits{Z.}},
\bauthor{\bsnm{Yang}, \binits{H.}},
\bauthor{\bsnm{Zhang}, \binits{J.}},
\bauthor{\bsnm{Chen}, \binits{Z.}},
\bauthor{\bsnm{Tang}, \binits{J.}},
\bauthor{\bsnm{Chen}, \binits{X.}},
\bauthor{\bsnm{Lin}, \binits{Y.}}, \betal:
\batitle{A survey on large language model based autonomous agents}.
\bjtitle{Frontiers of Computer Science}
\bvolume{18}(\bissue{6}),
\bfpage{186345}
(\byear{2024})
\end{barticle}
\endbibitem

\bibitem[\protect\citeauthoryear{Torkamaan et~al.}{2024}]{torkamaan2024challenges}
\begin{botherref}
\oauthor{\bsnm{Torkamaan}, \binits{H.}},
\oauthor{\bsnm{Steinert}, \binits{S.}},
\oauthor{\bsnm{Pera}, \binits{M.S.}},
\oauthor{\bsnm{Kudina}, \binits{O.}},
\oauthor{\bsnm{Freire}, \binits{S.K.}},
\oauthor{\bsnm{Verma}, \binits{H.}},
\oauthor{\bsnm{Kelly}, \binits{S.}},
\oauthor{\bsnm{Sekwenz}, \binits{M.-T.}},
\oauthor{\bsnm{Yang}, \binits{J.}},
\oauthor{\bsnm{Nunen}, \binits{K.}}, et al.:
Challenges and future directions for integration of large language models into socio-technical systems.
Behaviour \& Information Technology,
1--20
(2024)
\end{botherref}
\endbibitem

\bibitem[\protect\citeauthoryear{\"{O}zer et~al.}{2024}]{zer2024_1}
\begin{botherref}
\oauthor{\bsnm{\"{O}zer}, \binits{M.}},
\oauthor{\bsnm{Perc}, \binits{M.}},
\oauthor{\bsnm{Suna}, \binits{H.}}:
Artificial intelligence bias and the amplification of inequalities in the labor market.
Journal of Economy Culture and Society
\textbf{0}(0)
(2024)
\doiurl{10.26650/jecs2023-1415085}
\end{botherref}
\endbibitem

\bibitem[\protect\citeauthoryear{Binz et~al.}{2025}]{binz2025foundation}
\begin{botherref}
\oauthor{\bsnm{Binz}, \binits{M.}},
\oauthor{\bsnm{Akata}, \binits{E.}},
\oauthor{\bsnm{Bethge}, \binits{M.}},
\oauthor{\bsnm{Br{\"a}ndle}, \binits{F.}},
\oauthor{\bsnm{Callaway}, \binits{F.}},
\oauthor{\bsnm{Coda-Forno}, \binits{J.}},
\oauthor{\bsnm{Dayan}, \binits{P.}},
\oauthor{\bsnm{Demircan}, \binits{C.}},
\oauthor{\bsnm{Eckstein}, \binits{M.K.}},
\oauthor{\bsnm{{\'E}ltet{\H{o}}}, \binits{N.}}, et al.:
A foundation model to predict and capture human cognition.
Nature,
1--8
(2025)
\end{botherref}
\endbibitem

\bibitem[\protect\citeauthoryear{Qu and Wang}{2024}]{qu2024performance}
\begin{barticle}
\bauthor{\bsnm{Qu}, \binits{Y.}},
\bauthor{\bsnm{Wang}, \binits{J.}}:
\batitle{Performance and biases of large language models in public opinion simulation}.
\bjtitle{Humanities and Social Sciences Communications}
\bvolume{11}(\bissue{1}),
\bfpage{1}--\blpage{13}
(\byear{2024})
\end{barticle}
\endbibitem

\bibitem[\protect\citeauthoryear{Stureborg et~al.}{2024}]{stureborg2024large}
\begin{botherref}
\oauthor{\bsnm{Stureborg}, \binits{R.}},
\oauthor{\bsnm{Alikaniotis}, \binits{D.}},
\oauthor{\bsnm{Suhara}, \binits{Y.}}:
Large language models are inconsistent and biased evaluators.
arXiv preprint arXiv:2405.01724
(2024)
\end{botherref}
\endbibitem

\bibitem[\protect\citeauthoryear{Cinelli et~al.}{2021}]{cinelli2021echo}
\begin{barticle}
\bauthor{\bsnm{Cinelli}, \binits{M.}},
\bauthor{\bsnm{De~Francisci~Morales}, \binits{G.}},
\bauthor{\bsnm{Galeazzi}, \binits{A.}},
\bauthor{\bsnm{Quattrociocchi}, \binits{W.}},
\bauthor{\bsnm{Starnini}, \binits{M.}}:
\batitle{The echo chamber effect on social media}.
\bjtitle{Proceedings of the national academy of sciences}
\bvolume{118}(\bissue{9}),
\bfpage{2023301118}
(\byear{2021})
\end{barticle}
\endbibitem

\bibitem[\protect\citeauthoryear{Holton et~al.}{2015}]{holton2015reciprocity}
\begin{barticle}
\bauthor{\bsnm{Holton}, \binits{A.E.}},
\bauthor{\bsnm{Coddington}, \binits{M.}},
\bauthor{\bsnm{Lewis}, \binits{S.C.}},
\bauthor{\bsnm{De~Zuniga}, \binits{H.G.}}:
\batitle{Reciprocity and the news: The role of personal and social media reciprocity in news creation and consumption}.
\bjtitle{International journal of communication}
\bvolume{9},
\bfpage{22}
(\byear{2015})
\end{barticle}
\endbibitem

\bibitem[\protect\citeauthoryear{Khan}{2017}]{khan2017social}
\begin{barticle}
\bauthor{\bsnm{Khan}, \binits{M.L.}}:
\batitle{Social media engagement: What motivates user participation and consumption on youtube?}
\bjtitle{Computers in human behavior}
\bvolume{66},
\bfpage{236}--\blpage{247}
(\byear{2017})
\end{barticle}
\endbibitem

\bibitem[\protect\citeauthoryear{Avalle et~al.}{2024}]{avalle2024persistent}
\begin{barticle}
\bauthor{\bsnm{Avalle}, \binits{M.}},
\bauthor{\bsnm{Di~Marco}, \binits{N.}},
\bauthor{\bsnm{Etta}, \binits{G.}},
\bauthor{\bsnm{Sangiorgio}, \binits{E.}},
\bauthor{\bsnm{Alipour}, \binits{S.}},
\bauthor{\bsnm{Bonetti}, \binits{A.}},
\bauthor{\bsnm{Alvisi}, \binits{L.}},
\bauthor{\bsnm{Scala}, \binits{A.}},
\bauthor{\bsnm{Baronchelli}, \binits{A.}},
\bauthor{\bsnm{Cinelli}, \binits{M.}}, \betal:
\batitle{Persistent interaction patterns across social media platforms and over time}.
\bjtitle{Nature}
\bvolume{628}(\bissue{8008}),
\bfpage{582}--\blpage{589}
(\byear{2024})
\end{barticle}
\endbibitem

\bibitem[\protect\citeauthoryear{Kubin and Von~Sikorski}{2021}]{kubin2021role}
\begin{barticle}
\bauthor{\bsnm{Kubin}, \binits{E.}},
\bauthor{\bsnm{Von~Sikorski}, \binits{C.}}:
\batitle{The role of (social) media in political polarization: a systematic review}.
\bjtitle{Annals of the International Communication Association}
\bvolume{45}(\bissue{3}),
\bfpage{188}--\blpage{206}
(\byear{2021})
\end{barticle}
\endbibitem

\bibitem[\protect\citeauthoryear{Budak et~al.}{2024}]{budak2024misunderstanding}
\begin{barticle}
\bauthor{\bsnm{Budak}, \binits{C.}},
\bauthor{\bsnm{Nyhan}, \binits{B.}},
\bauthor{\bsnm{Rothschild}, \binits{D.M.}},
\bauthor{\bsnm{Thorson}, \binits{E.}},
\bauthor{\bsnm{Watts}, \binits{D.J.}}:
\batitle{Misunderstanding the harms of online misinformation}.
\bjtitle{Nature}
\bvolume{630}(\bissue{8015}),
\bfpage{45}--\blpage{53}
(\byear{2024})
\end{barticle}
\endbibitem

\bibitem[\protect\citeauthoryear{Lazer et~al.}{2018}]{lazer2018science}
\begin{barticle}
\bauthor{\bsnm{Lazer}, \binits{D.M.}},
\bauthor{\bsnm{Baum}, \binits{M.A.}},
\bauthor{\bsnm{Benkler}, \binits{Y.}},
\bauthor{\bsnm{Berinsky}, \binits{A.J.}},
\bauthor{\bsnm{Greenhill}, \binits{K.M.}},
\bauthor{\bsnm{Menczer}, \binits{F.}},
\bauthor{\bsnm{Metzger}, \binits{M.J.}},
\bauthor{\bsnm{Nyhan}, \binits{B.}},
\bauthor{\bsnm{Pennycook}, \binits{G.}},
\bauthor{\bsnm{Rothschild}, \binits{D.}}, \betal:
\batitle{The science of fake news}.
\bjtitle{Science}
\bvolume{359}(\bissue{6380}),
\bfpage{1094}--\blpage{1096}
(\byear{2018})
\end{barticle}
\endbibitem

\bibitem[\protect\citeauthoryear{Del~Vicario et~al.}{2016}]{del2016spreading}
\begin{barticle}
\bauthor{\bsnm{Del~Vicario}, \binits{M.}},
\bauthor{\bsnm{Bessi}, \binits{A.}},
\bauthor{\bsnm{Zollo}, \binits{F.}},
\bauthor{\bsnm{Petroni}, \binits{F.}},
\bauthor{\bsnm{Scala}, \binits{A.}},
\bauthor{\bsnm{Caldarelli}, \binits{G.}},
\bauthor{\bsnm{Stanley}, \binits{H.E.}},
\bauthor{\bsnm{Quattrociocchi}, \binits{W.}}:
\batitle{The spreading of misinformation online}.
\bjtitle{Proceedings of the national academy of Sciences}
\bvolume{113}(\bissue{3}),
\bfpage{554}--\blpage{559}
(\byear{2016})
\end{barticle}
\endbibitem

\bibitem[\protect\citeauthoryear{Gallup}{2018}]{gallup2018indicators}
\begin{botherref}
\oauthor{\bsnm{Gallup}, \binits{K.}}:
Indicators of news media trust.
John S. and James L. Knight Foundation Miami
(2018)
\end{botherref}
\endbibitem

\bibitem[\protect\citeauthoryear{Newman et~al.}{2018}]{newman2018digital}
\begin{botherref}
\oauthor{\bsnm{Newman}, \binits{N.}},
\oauthor{\bsnm{Fletcher}, \binits{R.}},
\oauthor{\bsnm{Levy}, \binits{D.}},
\oauthor{\bsnm{Nielsen}, \binits{R.}}:
The Reuters Institute Digital News Report 2018
\end{botherref}
\endbibitem

\bibitem[\protect\citeauthoryear{Bail et~al.}{2018}]{bail2018exposure}
\begin{barticle}
\bauthor{\bsnm{Bail}, \binits{C.A.}},
\bauthor{\bsnm{Argyle}, \binits{L.P.}},
\bauthor{\bsnm{Brown}, \binits{T.W.}},
\bauthor{\bsnm{Bumpus}, \binits{J.P.}},
\bauthor{\bsnm{Chen}, \binits{H.}},
\bauthor{\bsnm{Hunzaker}, \binits{M.F.}},
\bauthor{\bsnm{Lee}, \binits{J.}},
\bauthor{\bsnm{Mann}, \binits{M.}},
\bauthor{\bsnm{Merhout}, \binits{F.}},
\bauthor{\bsnm{Volfovsky}, \binits{A.}}:
\batitle{Exposure to opposing views on social media can increase political polarization}.
\bjtitle{Proceedings of the National Academy of Sciences}
\bvolume{115}(\bissue{37}),
\bfpage{9216}--\blpage{9221}
(\byear{2018})
\end{barticle}
\endbibitem

\bibitem[\protect\citeauthoryear{Falkenberg et~al.}{2022}]{falkenberg2022growing}
\begin{barticle}
\bauthor{\bsnm{Falkenberg}, \binits{M.}},
\bauthor{\bsnm{Galeazzi}, \binits{A.}},
\bauthor{\bsnm{Torricelli}, \binits{M.}},
\bauthor{\bsnm{Di~Marco}, \binits{N.}},
\bauthor{\bsnm{Larosa}, \binits{F.}},
\bauthor{\bsnm{Sas}, \binits{M.}},
\bauthor{\bsnm{Mekacher}, \binits{A.}},
\bauthor{\bsnm{Pearce}, \binits{W.}},
\bauthor{\bsnm{Zollo}, \binits{F.}},
\bauthor{\bsnm{Quattrociocchi}, \binits{W.}}, \betal:
\batitle{Growing polarization around climate change on social media}.
\bjtitle{Nature Climate Change}
\bvolume{12}(\bissue{12}),
\bfpage{1114}--\blpage{1121}
(\byear{2022})
\end{barticle}
\endbibitem

\bibitem[\protect\citeauthoryear{Cinelli et~al.}{2020}]{cinelli2020covid}
\begin{barticle}
\bauthor{\bsnm{Cinelli}, \binits{M.}},
\bauthor{\bsnm{Quattrociocchi}, \binits{W.}},
\bauthor{\bsnm{Galeazzi}, \binits{A.}},
\bauthor{\bsnm{Valensise}, \binits{C.M.}},
\bauthor{\bsnm{Brugnoli}, \binits{E.}},
\bauthor{\bsnm{Schmidt}, \binits{A.L.}},
\bauthor{\bsnm{Zola}, \binits{P.}},
\bauthor{\bsnm{Zollo}, \binits{F.}},
\bauthor{\bsnm{Scala}, \binits{A.}}:
\batitle{The covid-19 social media infodemic}.
\bjtitle{Scientific reports}
\bvolume{10}(\bissue{1}),
\bfpage{1}--\blpage{10}
(\byear{2020})
\end{barticle}
\endbibitem

\bibitem[\protect\citeauthoryear{Metzger and Flanagin}{2015}]{metzger2015psychological}
\begin{botherref}
\oauthor{\bsnm{Metzger}, \binits{M.J.}},
\oauthor{\bsnm{Flanagin}, \binits{A.J.}}:
Psychological approaches to credibility assessment online.
The handbook of the psychology of communication technology,
445--466
(2015)
\end{botherref}
\endbibitem

\bibitem[\protect\citeauthoryear{Rieh}{2010}]{rieh2010credibility}
\begin{barticle}
\bauthor{\bsnm{Rieh}, \binits{S.Y.}}:
\batitle{Credibility and cognitive authority of information}.
\bjtitle{Encyclopedia of library and information sciences}
\bvolume{1}(\bissue{1}),
\bfpage{1337}--\blpage{1344}
(\byear{2010})
\end{barticle}
\endbibitem

\bibitem[\protect\citeauthoryear{Metzger et~al.}{2010}]{metzger2010social}
\begin{barticle}
\bauthor{\bsnm{Metzger}, \binits{M.J.}},
\bauthor{\bsnm{Flanagin}, \binits{A.J.}},
\bauthor{\bsnm{Medders}, \binits{R.B.}}:
\batitle{Social and heuristic approaches to credibility evaluation online}.
\bjtitle{Journal of communication}
\bvolume{60}(\bissue{3}),
\bfpage{413}--\blpage{439}
(\byear{2010})
\end{barticle}
\endbibitem

\bibitem[\protect\citeauthoryear{L{\"u}hring et~al.}{2025}]{luhring2025best}
\begin{botherref}
\oauthor{\bsnm{L{\"u}hring}, \binits{J.}},
\oauthor{\bsnm{Metzler}, \binits{H.}},
\oauthor{\bsnm{Lazzaroni}, \binits{R.}},
\oauthor{\bsnm{Shetty}, \binits{A.}},
\oauthor{\bsnm{Lasser}, \binits{J.}}:
Best practices for source-based research on misinformation and news trustworthiness using newsguard.
Journal of Quantitative Description: Digital Media
\textbf{5}
(2025)
\end{botherref}
\endbibitem

\bibitem[\protect\citeauthoryear{Nakov et~al.}{2021}]{nakov2021survey}
\begin{botherref}
\oauthor{\bsnm{Nakov}, \binits{P.}},
\oauthor{\bsnm{Sencar}, \binits{H.T.}},
\oauthor{\bsnm{An}, \binits{J.}},
\oauthor{\bsnm{Kwak}, \binits{H.}}:
A survey on predicting the factuality and the bias of news media.
arXiv preprint arXiv:2103.12506
(2021)
\end{botherref}
\endbibitem

\bibitem[\protect\citeauthoryear{{NewsGuard Technologies}}{}]{newsguard2024}
\begin{botherref}
\oauthor{\bsnm{{NewsGuard Technologies}}}:
Rating Process and Criteria.
\url{https://www.newsguardtech.com/ratings/rating\%20-process-\%20criteria/}.
Accessed: 2024-11-26
\end{botherref}
\endbibitem

\bibitem[\protect\citeauthoryear{OpenAI}{2024}]{openai2024}
\begin{botherref}
\oauthor{\bsnm{OpenAI}}:
GPT-4o Technical Report
(2024).
\url{https://openai.com/index/gpt-4o-mini-advancing-cost-efficient-intelligence/}
\end{botherref}
\endbibitem

\bibitem[\protect\citeauthoryear{Team et~al.}{2024}]{team2024gemini}
\begin{botherref}
\oauthor{\bsnm{Team}, \binits{G.}},
\oauthor{\bsnm{Georgiev}, \binits{P.}},
\oauthor{\bsnm{Lei}, \binits{V.I.}},
\oauthor{\bsnm{Burnell}, \binits{R.}},
\oauthor{\bsnm{Bai}, \binits{L.}},
\oauthor{\bsnm{Gulati}, \binits{A.}},
\oauthor{\bsnm{Tanzer}, \binits{G.}},
\oauthor{\bsnm{Vincent}, \binits{D.}},
\oauthor{\bsnm{Pan}, \binits{Z.}},
\oauthor{\bsnm{Wang}, \binits{S.}}, et al.:
Gemini 1.5: Unlocking multimodal understanding across millions of tokens of context.
arXiv preprint arXiv:2403.05530
(2024)
\end{botherref}
\endbibitem

\bibitem[\protect\citeauthoryear{AI}{2024}]{meta2024}
\begin{botherref}
\oauthor{\bsnm{AI}, \binits{M.}}:
LLaMA 3.1: Advancements in Open-Weight LLMs
(2024).
\url{https://ai.meta.com/blog/meta-llama-3-1/}
\end{botherref}
\endbibitem

\bibitem[\protect\citeauthoryear{T{\"o}rnberg}{2024}]{tornberg2024large}
\begin{botherref}
\oauthor{\bsnm{T{\"o}rnberg}, \binits{P.}}:
Large language models outperform expert coders and supervised classifiers at annotating political social media messages.
Social Science Computer Review,
08944393241286471
(2024)
\end{botherref}
\endbibitem

\bibitem[\protect\citeauthoryear{Gilardi et~al.}{2023}]{gilardi2023chatgpt}
\begin{barticle}
\bauthor{\bsnm{Gilardi}, \binits{F.}},
\bauthor{\bsnm{Alizadeh}, \binits{M.}},
\bauthor{\bsnm{Kubli}, \binits{M.}}:
\batitle{Chatgpt outperforms crowd workers for text-annotation tasks}.
\bjtitle{Proceedings of the National Academy of Sciences}
\bvolume{120}(\bissue{30}),
\bfpage{2305016120}
(\byear{2023})
\end{barticle}
\endbibitem

\bibitem[\protect\citeauthoryear{Wu et~al.}{2023}]{wu2023large}
\begin{botherref}
\oauthor{\bsnm{Wu}, \binits{P.Y.}},
\oauthor{\bsnm{Nagler}, \binits{J.}},
\oauthor{\bsnm{Tucker}, \binits{J.A.}},
\oauthor{\bsnm{Messing}, \binits{S.}}:
Large language models can be used to estimate the latent positions of politicians.
arXiv preprint arXiv:2303.12057
(2023)
\end{botherref}
\endbibitem

\bibitem[\protect\citeauthoryear{Chiang and Lee}{2023}]{chiang2023can}
\begin{botherref}
\oauthor{\bsnm{Chiang}, \binits{C.-H.}},
\oauthor{\bsnm{Lee}, \binits{H.-y.}}:
Can large language models be an alternative to human evaluations?
arXiv preprint arXiv:2305.01937
(2023)
\end{botherref}
\endbibitem

\bibitem[\protect\citeauthoryear{Krugmann and Hartmann}{2024}]{krugmann2024sentiment}
\begin{barticle}
\bauthor{\bsnm{Krugmann}, \binits{J.O.}},
\bauthor{\bsnm{Hartmann}, \binits{J.}}:
\batitle{Sentiment analysis in the age of generative ai}.
\bjtitle{Customer Needs and Solutions}
\bvolume{11}(\bissue{1}),
\bfpage{3}
(\byear{2024})
\end{barticle}
\endbibitem

\bibitem[\protect\citeauthoryear{Yang and Menczer}{2023}]{yang2023large}
\begin{botherref}
\oauthor{\bsnm{Yang}, \binits{K.-C.}},
\oauthor{\bsnm{Menczer}, \binits{F.}}:
Large language models can rate news outlet credibility.
arXiv preprint arXiv:2304.00228
(2023)
\end{botherref}
\endbibitem

\bibitem[\protect\citeauthoryear{Hoes et~al.}{2023}]{hoes2023leveraging}
\begin{botherref}
\oauthor{\bsnm{Hoes}, \binits{E.}},
\oauthor{\bsnm{Altay}, \binits{S.}},
\oauthor{\bsnm{Bermeo}, \binits{J.}}:
Leveraging chatgpt for efficient fact-checking.
PsyArXiv. April
\textbf{3}
(2023)
\end{botherref}
\endbibitem

\bibitem[\protect\citeauthoryear{Quelle and Bovet}{2024}]{quelle2024perils}
\begin{barticle}
\bauthor{\bsnm{Quelle}, \binits{D.}},
\bauthor{\bsnm{Bovet}, \binits{A.}}:
\batitle{The perils and promises of fact-checking with large language models}.
\bjtitle{Frontiers in Artificial Intelligence}
\bvolume{7},
\bfpage{1341697}
(\byear{2024})
\end{barticle}
\endbibitem

\bibitem[\protect\citeauthoryear{Hernandes and Corsi}{2024}]{hernandes2024llms}
\begin{botherref}
\oauthor{\bsnm{Hernandes}, \binits{R.}},
\oauthor{\bsnm{Corsi}, \binits{G.}}:
Llms left, right, and center: Assessing gpt's capabilities to label political bias from web domains.
arXiv preprint arXiv:2407.14344
(2024)
\end{botherref}
\endbibitem

\bibitem[\protect\citeauthoryear{Hu et~al.}{2024}]{hu2024generative}
\begin{botherref}
\oauthor{\bsnm{Hu}, \binits{T.}},
\oauthor{\bsnm{Kyrychenko}, \binits{Y.}},
\oauthor{\bsnm{Rathje}, \binits{S.}},
\oauthor{\bsnm{Collier}, \binits{N.}},
\oauthor{\bsnm{Linden}, \binits{S.}},
\oauthor{\bsnm{Roozenbeek}, \binits{J.}}:
Generative language models exhibit social identity biases.
Nature Computational Science,
1--11
(2024)
\end{botherref}
\endbibitem

\bibitem[\protect\citeauthoryear{Motoki et~al.}{2025}]{motoki2025assessing}
\begin{botherref}
\oauthor{\bsnm{Motoki}, \binits{F.Y.}},
\oauthor{\bsnm{Neto}, \binits{V.P.}},
\oauthor{\bsnm{Rangel}, \binits{V.}}:
Assessing political bias and value misalignment in generative artificial intelligence.
Journal of Economic Behavior \& Organization,
106904
(2025)
\end{botherref}
\endbibitem

\bibitem[\protect\citeauthoryear{Strachan et~al.}{2024}]{strachan2024testing}
\begin{botherref}
\oauthor{\bsnm{Strachan}, \binits{J.W.}},
\oauthor{\bsnm{Albergo}, \binits{D.}},
\oauthor{\bsnm{Borghini}, \binits{G.}},
\oauthor{\bsnm{Pansardi}, \binits{O.}},
\oauthor{\bsnm{Scaliti}, \binits{E.}},
\oauthor{\bsnm{Gupta}, \binits{S.}},
\oauthor{\bsnm{Saxena}, \binits{K.}},
\oauthor{\bsnm{Rufo}, \binits{A.}},
\oauthor{\bsnm{Panzeri}, \binits{S.}},
\oauthor{\bsnm{Manzi}, \binits{G.}}, et al.:
Testing theory of mind in large language models and humans.
Nature Human Behaviour,
1--11
(2024)
\end{botherref}
\endbibitem

\bibitem[\protect\citeauthoryear{Coppolillo et~al.}{2025}]{coppolillo2025unmasking}
\begin{botherref}
\oauthor{\bsnm{Coppolillo}, \binits{E.}},
\oauthor{\bsnm{Manco}, \binits{G.}},
\oauthor{\bsnm{Aiello}, \binits{L.M.}}:
Unmasking conversational bias in ai multiagent systems.
arXiv preprint arXiv:2501.14844
(2025)
\end{botherref}
\endbibitem

\bibitem[\protect\citeauthoryear{Safdari et~al.}{2023}]{safdari2023personality}
\begin{botherref}
\oauthor{\bsnm{Safdari}, \binits{M.}},
\oauthor{\bsnm{Serapio-Garc{\'\i}a}, \binits{G.}},
\oauthor{\bsnm{Crepy}, \binits{C.}},
\oauthor{\bsnm{Fitz}, \binits{S.}},
\oauthor{\bsnm{Romero}, \binits{P.}},
\oauthor{\bsnm{Sun}, \binits{L.}},
\oauthor{\bsnm{Abdulhai}, \binits{M.}},
\oauthor{\bsnm{Faust}, \binits{A.}},
\oauthor{\bsnm{Matari{\'c}}, \binits{M.}}:
Personality traits in large language models.
arXiv preprint arXiv:2307.00184
(2023)
\end{botherref}
\endbibitem

\bibitem[\protect\citeauthoryear{Di~Marco et~al.}{2024}]{di2024patterns}
\begin{barticle}
\bauthor{\bsnm{Di~Marco}, \binits{N.}},
\bauthor{\bsnm{Loru}, \binits{E.}},
\bauthor{\bsnm{Bonetti}, \binits{A.}},
\bauthor{\bsnm{Serra}, \binits{A.O.G.}},
\bauthor{\bsnm{Cinelli}, \binits{M.}},
\bauthor{\bsnm{Quattrociocchi}, \binits{W.}}:
\batitle{Patterns of linguistic simplification on social media platforms over time}.
\bjtitle{Proceedings of the National Academy of Sciences}
\bvolume{121}(\bissue{50}),
\bfpage{2412105121}
(\byear{2024})
\end{barticle}
\endbibitem

\bibitem[\protect\citeauthoryear{Pf{\"a}nder and Altay}{2025}]{pfander2025spotting}
\begin{botherref}
\oauthor{\bsnm{Pf{\"a}nder}, \binits{J.}},
\oauthor{\bsnm{Altay}, \binits{S.}}:
Spotting false news and doubting true news: a systematic review and meta-analysis of news judgements.
Nature human behaviour,
1--12
(2025)
\end{botherref}
\endbibitem

\bibitem[\protect\citeauthoryear{Pennycook and Rand}{2019}]{pennycook2019lazy}
\begin{barticle}
\bauthor{\bsnm{Pennycook}, \binits{G.}},
\bauthor{\bsnm{Rand}, \binits{D.G.}}:
\batitle{Lazy, not biased: Susceptibility to partisan fake news is better explained by lack of reasoning than by motivated reasoning}.
\bjtitle{Cognition}
\bvolume{188},
\bfpage{39}--\blpage{50}
(\byear{2019})
\end{barticle}
\endbibitem

\bibitem[\protect\citeauthoryear{Wineburg and McGrew}{2019}]{wineburg2019lateral}
\begin{barticle}
\bauthor{\bsnm{Wineburg}, \binits{S.}},
\bauthor{\bsnm{McGrew}, \binits{S.}}:
\batitle{Lateral reading and the nature of expertise: Reading less and learning more when evaluating digital information}.
\bjtitle{Teachers College Record}
\bvolume{121}(\bissue{11}),
\bfpage{1}--\blpage{40}
(\byear{2019})
\end{barticle}
\endbibitem

\bibitem[\protect\citeauthoryear{Fiske}{2018}]{fiske2018social}
\begin{bbook}
\bauthor{\bsnm{Fiske}, \binits{S.}}:
\bbtitle{Social Cognition: Selected Works of Susan Fiske}.
\bpublisher{Routledge}, \blocation{???}
(\byear{2018})
\end{bbook}
\endbibitem

\bibitem[\protect\citeauthoryear{Tversky and Kahneman}{1974}]{tversky1974judgment}
\begin{barticle}
\bauthor{\bsnm{Tversky}, \binits{A.}},
\bauthor{\bsnm{Kahneman}, \binits{D.}}:
\batitle{Judgment under uncertainty: Heuristics and biases: Biases in judgments reveal some heuristics of thinking under uncertainty.}
\bjtitle{science}
\bvolume{185}(\bissue{4157}),
\bfpage{1124}--\blpage{1131}
(\byear{1974})
\end{barticle}
\endbibitem

\bibitem[\protect\citeauthoryear{Reber and Schwarz}{1999}]{reber1999effects}
\begin{barticle}
\bauthor{\bsnm{Reber}, \binits{R.}},
\bauthor{\bsnm{Schwarz}, \binits{N.}}:
\batitle{Effects of perceptual fluency on judgments of truth}.
\bjtitle{Consciousness and cognition}
\bvolume{8}(\bissue{3}),
\bfpage{338}--\blpage{342}
(\byear{1999})
\end{barticle}
\endbibitem

\bibitem[\protect\citeauthoryear{Fazio et~al.}{2015}]{fazio2015knowledge}
\begin{barticle}
\bauthor{\bsnm{Fazio}, \binits{L.K.}},
\bauthor{\bsnm{Brashier}, \binits{N.M.}},
\bauthor{\bsnm{Payne}, \binits{B.K.}},
\bauthor{\bsnm{Marsh}, \binits{E.J.}}:
\batitle{Knowledge does not protect against illusory truth.}
\bjtitle{Journal of experimental psychology: general}
\bvolume{144}(\bissue{5}),
\bfpage{993}
(\byear{2015})
\end{barticle}
\endbibitem

\bibitem[\protect\citeauthoryear{Kahneman}{2003}]{kahneman2002maps}
\begin{barticle}
\bauthor{\bsnm{Kahneman}, \binits{D.}}:
\batitle{Maps of bounded rationality: Psychology for behavioral economics}.
\bjtitle{American economic review}
\bvolume{93}(\bissue{5}),
\bfpage{1449}--\blpage{1475}
(\byear{2003})
\end{barticle}
\endbibitem

\bibitem[\protect\citeauthoryear{Taber and Lodge}{2006}]{taber2006motivated}
\begin{barticle}
\bauthor{\bsnm{Taber}, \binits{C.S.}},
\bauthor{\bsnm{Lodge}, \binits{M.}}:
\batitle{Motivated skepticism in the evaluation of political beliefs}.
\bjtitle{American journal of political science}
\bvolume{50}(\bissue{3}),
\bfpage{755}--\blpage{769}
(\byear{2006})
\end{barticle}
\endbibitem

\bibitem[\protect\citeauthoryear{}{2024}]{githubrequests}
\begin{botherref}
{G}it{H}ub - psf/requests: {A} simple, yet elegant, {H}{T}{T}{P} library. --- github.com.
\url{https://github.com/psf/requests}
(2024)
\end{botherref}
\endbibitem

\bibitem[\protect\citeauthoryear{Richardson}{2024}]{crummyBeautifulSoup}
\begin{botherref}
\oauthor{\bsnm{Richardson}, \binits{L.}}:
{B}eautiful {S}oup: {W}e called him {T}ortoise because he taught us. --- crummy.com.
\url{https://www.crummy.com/software/BeautifulSoup/}
(2024)
\end{botherref}
\endbibitem

\bibitem[\protect\citeauthoryear{}{2025}]{githubADK}
\begin{botherref}
{G}it{H}ub - google/adk-python: {A}n open-source, code-first {P}ython toolkit for building, evaluating, and deploying sophisticated {A}{I} agents with flexibility and control. --- github.com.
\url{https://github.com/google/adk-python}
(2025)
\end{botherref}
\endbibitem

\bibitem[\protect\citeauthoryear{Wang et~al.}{2024}]{wang2024code}
\begin{bchapter}
\bauthor{\bsnm{Wang}, \binits{X.}},
\bauthor{\bsnm{Chen}, \binits{Y.}},
\bauthor{\bsnm{Yuan}, \binits{L.}},
\bauthor{\bsnm{Zhang}, \binits{Y.}},
\bauthor{\bsnm{Li}, \binits{Y.}},
\bauthor{\bsnm{Peng}, \binits{H.}},
\bauthor{\bsnm{Ji}, \binits{H.}}:
\bctitle{Executable code actions elicit better {LLM} agents}.
In: \beditor{\bsnm{Salakhutdinov}, \binits{R.}},
\beditor{\bsnm{Kolter}, \binits{Z.}},
\beditor{\bsnm{Heller}, \binits{K.}},
\beditor{\bsnm{Weller}, \binits{A.}},
\beditor{\bsnm{Oliver}, \binits{N.}},
\beditor{\bsnm{Scarlett}, \binits{J.}},
\beditor{\bsnm{Berkenkamp}, \binits{F.}} (eds.)
\bbtitle{Proceedings of the 41st International Conference on Machine Learning}.
\bsertitle{Proceedings of Machine Learning Research},
vol. \bseriesno{235},
pp. \bfpage{50208}--\blpage{50232}.
\bpublisher{PMLR}, \blocation{???}
(\byear{2024})
\end{bchapter}
\endbibitem

\bibitem[\protect\citeauthoryear{}{2025}]{githubMarkitdown}
\begin{botherref}
{G}it{H}ub - microsoft/markitdown: {P}ython tool for converting files and office documents to {M}arkdown. --- github.com.
\url{https://github.com/microsoft/markitdown}
(2025)
\end{botherref}
\endbibitem

\end{thebibliography}
\end{document}